\documentclass{article}
\usepackage[preprint]{neurips_2026}
\usepackage{amsmath,amssymb}
\usepackage{graphicx}
\usepackage{microtype}
\usepackage{xcolor}
\usepackage{hyperref}
\usepackage{tikz}
\usetikzlibrary{arrows.meta,positioning,backgrounds}

\definecolor{outlineblue}{RGB}{0,82,155}

\title{Differentiable Electricity-Market Clearing for Gradient-Based Planning}
\author{ Luca Mungo\textsuperscript{1,2} \\ \texttt{luca@mcsm.ai} \\ \And Maarten P. Scholl\textsuperscript{1} \\ \texttt{maarten@mcsm.ai} \And Arnau Quera-Bofarull\textsuperscript{1} \\ \texttt{arnau@mcsm.ai} \\  \textsuperscript{1}Macrocosm Inc.\quad \textsuperscript{2}Institute for New Economic Thinking, University of Oxford}

\begin{document}
\maketitle

\begin{abstract}
Planning a large data center is difficult because a facility big enough to matter changes the electricity prices it will pay. Those prices are set by market
clearing, a constrained optimization problem solved anew in every operating
condition. However, simulating the market tells a planner how a candidate plan performs but not how to improve it. Here we treat market clearing as a
differentiable optimization layer: each forward pass solves the market, and
reverse-mode automatic differentiation propagates the planning cost back through the
cleared prices to the plan. After validating these gradients against finite
differences, we apply them to a concrete problem: allocating 50 MW of
data-center load across six candidate buses in two synthetic networks, under a fixed cost per active site, evaluated over 36 operating states. Judged against exhaustive enumeration of all site combinations, gradient
optimization recovers the continuous allocations almost exactly, with
worst-case objective gaps of 2.3\% and 8.5\% of the cost difference between
the best and worst single site. Its one systematic error is instructive: near the costs at which a site should close, the smooth relaxation of the discrete site count shrinks the site rather than closing it, so discrete transitions arrive late. Differentiable market clearing thus turns market-aware planning into a problem gradients can search.
\end{abstract}

\section{Introduction}

Planning investments in electricity markets is challenging because investment decisions can substantially alter dispatch, congestion, and electricity prices. These outcomes are determined by market clearing, an optimization problem that minimizes generation costs subject to system constraints and, in doing so, determines generator dispatch and electricity prices~\citep{kirschen2004}. Evaluating a candidate investment therefore requires clearing the market under that investment, often across many operating conditions. This process tells the planner how a proposed investment will perform, but not how the investment should be adjusted, so finding the best plan may require searching a large space of alternatives. We study this problem in a simple setting: how should a planner distribute 50 MW of new data-center load across six candidate buses when each allocation changes the prices the data centers will pay?

We ask whether the gradient of the planning objective with respect to the investment decision can guide the search over candidate plans. Obtaining these sensitivities requires differentiating through the market-clearing problem itself. The planning objective depends on nodal prices, which are dual outputs of a constrained optimization problem, so its gradient must account for how those prices respond to changes in the plan. Optimization layers and automatic implicit differentiation provide the computational tools for obtaining these gradients \citep{amos2017,agrawal2019layers,blondel2022,besancon2024}. This yields a differentiable market-clearing model that can be embedded directly within a gradient-based planning procedure.

Each candidate plan is evaluated by clearing the market across its operating states, and reverse-mode automatic differentiation propagates the resulting cost through nodal prices to the planning variables; electricity-market clearing becomes a differentiable optimization layer. We use this approach to allocate a load across candidate buses in two families of synthetic networks. Because site selection is discrete, we optimize a smooth relaxation of the site-count penalty, then commit and refine the resulting plans.

Across both fixed-cost sweeps, the gradient-based procedure produces plans whose objective values closely track the numerical optimum defined in Appendix~\ref{app:planning-details}. This pattern persists across five near-uniform initializations, with the largest departures from the numerical optimum and the greatest variation occurring near portfolio switches. Although we evaluate only one instance of each network family, these case studies show that differentiable market clearing can guide an end-to-end planning procedure toward near-optimal solutions, while also exposing the difficulty of smooth relaxations near discrete portfolio switches.

Our approach builds on two established lines of work. Sensitivity analysis of optimal power flow has long been used to characterize how locational marginal prices respond to changes in demand and other market parameters \citep{conejo2005,yang2014}, while differentiable optimization layers provide general tools for propagating derivatives through constrained optimization problems \citep{amos2017,agrawal2019layers,blondel2022,besancon2024}. Data-center siting has likewise been formulated as a bilevel formulation (a nested optimization problem in which location decisions affect subsequent electricity-market outcomes) \citep{zeng2024} but has typically been addressed using reformulation and decomposition techniques. While small bilevel problems can be addressed through single-level reformulations, evidence from large-scale electricity-grid planning suggests that implicit-gradient methods can outperform reformulation-based approaches as the problem size and number of operating scenarios grow~\citep{degleris2026}. This motivates studying gradient-based methods for market-aware data-center planning. Our setting combines these threads by propagating derivatives through cleared nodal prices to optimize the spatial allocation of a large new load. Unlike continuous capacity-expansion settings, the resulting outer problem also contains a discrete site-selection decision; we address it through a smooth relaxation and evaluate the resulting plans against an enumerated numerical reference.

\section{Differentiable market-aware planning}

Let $x\in\mathcal X$ denote a decision available to the planner, let $\theta$
collect the fixed physical properties of the electricity system, and let
$\omega_s$ describe the demand and generation offers in operating state $s$.
Given these three inputs, the market clears by solving
\begin{equation}
  y_s^\star(x;\theta,\omega_s)
  \in\operatorname*{arg\,min}_{y_s\in\mathcal Y_s}
  C_s(y_s)\quad\text{subject to}\quad
  b_{s,i}(y_s,x)=0,\quad
  \left|f_{s,\ell}(y_s,x)\right|\leq\overline f_\ell.
  \label{eq:market-clearing}
\end{equation}
Here $y_s$ is the market dispatch, while $C_s$ and $\mathcal Y_s$, determined by $\omega_s$, specify the generation cost and available quantities. The constraints apply to every bus $i$ and line $\ell$. The function $b_{s,i}$ is the balance at bus $i$ (demand and exports minus local generation and imports) and $f_{s,\ell}$ is the flow on a line with capacity $\overline f_\ell$; these depend on the fixed system $\theta$. The dual variable of $b_{s,i}=0$ is the nodal price $\lambda_{s,i}(x;\theta,\omega_s)$; $\lambda_s$ collects these prices. Holding $\theta$ and the market states fixed, we write $\lambda_s(x)$ below. The planner solves
\begin{equation}
  \min_{x\in\mathcal X} J(x),\qquad
  J(x)=\sum_{s\in\mathcal S}
  \ell_s\!\left(x,\lambda_s(x)\right)+R(x),
  \label{eq:nested-planning}
\end{equation}
where $\ell_s$ is the state-specific market-dependent cost evaluated from the plan and the cleared nodal prices, with any state weights absorbed into $\ell_s$, and $R$ contains costs incurred directly by the plan. Evaluating $J(x)$ therefore requires solving the market-clearing problem in every operating state, and the dependence of $\lambda_s$ on $x$ captures the market response to the plan. In our application, $x=q$ is the allocation of additional nodal load, $\theta$ encodes the network and its line properties, and $\omega_s$ contains baseline demand and generation offers. The expenditure of the additional load in state $s$ is $q\cdot\lambda_s(q)$, whose gradient is
\begin{equation}
  \nabla_q\!\left[q\cdot\lambda_s(q)\right]
  =\lambda_s(q)+
  \left(\frac{\partial\lambda_s}{\partial q}\right)^{\!\top} q.
  \label{eq:market-mediated-gradient}
\end{equation}
The first term measures the effect of moving load while holding prices fixed. The second measures the effect of the resulting price changes on the whole portfolio. Figure~\ref{fig:flow} summarizes the forward market evaluation and reverse propagation of the planning gradient.

\definecolor{flowblue}{RGB}{8,48,107}
\definecolor{floworange}{RGB}{214,125,95}

\begin{figure}[ht]
  \centering
  \begin{tikzpicture}[
    >=Stealth,
    node distance=0.55cm,
    font=\footnotesize,
    box/.style={draw=black!60, fill=white, rounded corners=2pt,
      minimum height=0.62cm, inner xsep=6pt, inner ysep=3pt},
    market/.style={draw=flowblue, semithick, rounded corners=2pt,
      fill=flowblue!8, minimum height=0.62cm, inner xsep=7pt, inner ysep=3pt},
    param/.style={draw=black!35, fill=white, rounded corners=2pt,
      minimum height=0.5cm, inner xsep=5pt, inner ysep=2pt, text=black!70,
      font=\scriptsize},
    forward/.style={->, semithick},
    data/.style={->, thin, black!55},
    backward/.style={->, semithick, dashed, floworange}
  ]
    \node[box] (x) {$x$};
    \node[market, right=0.95cm of x, align=center] (clear)
      {$\arg\min_{y_s\in\mathcal Y_s} C_s(y_s)$\\[-1pt]
       {\scriptsize $\text{s.t. } b_{s,i}(y_s,x)=0,\;
        |f_{s,\ell}(y_s,x)|\le\overline f_\ell$}};
    \node[box, right=of clear] (state) {$y_s^\star,\ \lambda_s$};
    \node[box, right=of state] (loss) {$\ell_s(x,\lambda_s)$};
    \node[box, right=1.25cm of loss] (J) {$J(x)$};

    \node[param, above=0.22cm of clear, xshift=-0.95cm] (theta)
      {fixed system $\theta$};
    \node[param, above=0.22cm of clear, xshift=0.95cm] (omega)
      {market states $\omega_s$};

    \begin{scope}[on background layer]
      \foreach \offset in {4.4pt, 2.2pt} {
        \draw[draw=flowblue!55, fill=flowblue!4, rounded corners=2pt]
          ([shift={(\offset,-\offset)}]clear.south west) rectangle
          ([shift={(\offset,-\offset)}]clear.north east);
        \foreach \pernode in {omega, state, loss}
          \draw[draw=black!40, fill=white, rounded corners=2pt]
            ([shift={(\offset,-\offset)}]\pernode.south west) rectangle
            ([shift={(\offset,-\offset)}]\pernode.north east);
      }
    \end{scope}

    \draw[forward] (x) -- (clear);
    \draw[forward] ([xshift=4.4pt]clear.east) -- (state);
    \draw[forward] ([xshift=4.4pt]state.east) -- (loss);
    \draw[forward] ([xshift=4.4pt]loss.east) -- node[above, font=\scriptsize, xshift=4pt] {$\sum_s$}
      node[below, font=\scriptsize, xshift=1pt] {$+\,R(x)$} (J);
    \draw[data] (theta.south) -- (theta.south |- clear.north);
    \draw[data] (omega.south) -- (omega.south |- clear.north);

    \draw[backward] (J.south) .. controls +(0,-0.68) and +(0,-0.68) ..
      node[below, font=\scriptsize, yshift=-1pt]
        {reverse pass $\nabla_x J$, differentiated through the solved market}
      (x.south);
  \end{tikzpicture}
  \caption{Forward evaluation and reverse pass of the planning objective. The
  plan $x$ sets the inputs of every state's market-clearing problem, whose
  solved dispatch and nodal prices price the plan; the fixed system $\theta$
  and market state $\omega_s$ enter as data and are not differentiated. The
  state costs $\ell_s$, which depend on $x$ directly and through the prices,
  sum with the direct cost $R$ to $J(x)$; the dashed path propagates the
  planning gradient back through the solved clearing problem of every state.}
  \label{fig:flow}
\end{figure}
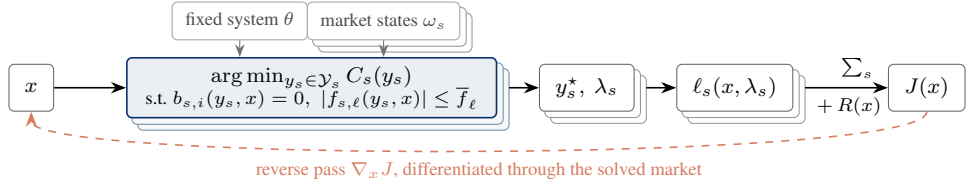

Advances in differentiable optimization make it possible to differentiate through the solutions of constrained optimization problems~\citep{amos2017,blondel2022}. We treat electricity-market clearing as a differentiable layer using \texttt{DiffOpt.jl}~\citep{besancon2024}: each forward pass solves the market, and reverse-mode differentiation propagates the planning loss through both its direct dependence on $x$ and its market-mediated dependence through nodal prices. We use the resulting gradients to minimize Equation~\eqref{eq:nested-planning} with a standard gradient-based optimizer.

\section{Data-center planning experiment}

\subsection{Experimental setup}

We consider the problem of building a portfolio of data centers to serve a total load of $D=50$ MW. We run the experiment on two synthetic network instances: a connected Erd\H{o}s--R\'enyi graph~\citep{newman2010networks} and a GeoDe graph~\citep{dey2023geode}. The Erd\H{o}s--R\'enyi model provides a nonspatial baseline in which connections are sampled at random, conditional on the network being connected. GeoDe instead uses a geometric construction with predominantly local connections to model the sparse, spatial structure of power grids. In each instance, we allow construction at six candidate buses and let $w_i$ denote the fraction of the total load assigned to candidate $i$; the allocation satisfies $w_i\geq0$ and $\sum_i w_i=1$, and the load at candidate $i$ is $q_i=Dw_i$. We imagine that the data centers will operate over a set $\mathcal{S}$ of 36 operating states. For state $s$, the additional load changes market clearing and hence the locational marginal prices $\lambda_s(q)$ faced by the portfolio.

We model a fixed construction cost $\kappa$ per active site. The physical planning objective is
\begin{equation}
  J(q;\kappa)
  = \frac{1}{|\mathcal{S}|}\sum_{s \in \mathcal{S}}q \cdot \lambda_s(q)
    + \kappa\lVert q\rVert_0.
  \label{eq:planning-objective}
\end{equation}
The first term is the mean nodal expenditure after re-clearing every state. The second term is a fixed charge for the number of sites receiving a positive allocation. Thus, the construction term adds $\kappa$ once for each active candidate site, irrespective of the load assigned to that site. In the reported physical evaluation, we apply this count after the 1 MW commitment rule described below. The exact site count jumps whenever a site opens or closes and therefore provides no useful gradient. During optimization we parameterize $w=\operatorname{softmax}(z)$ and replace the site count by $K_{\tau}(w)=\sum_i[1-\exp(-w_i/\tau)]$. The forward pass uses a sharp site-count approximation, while the straight-through backward pass uses a smoother temperature that is annealed during training. This relaxation affects only the construction-cost term; every market clear receives the full fractional load $q=Dw$.

For each $\kappa$, Adam produces five dense relaxed allocations from near-uniform initializations. We obtain physical solutions by discarding allocations below 1 MW, refining on the committed support, and re-clearing all 36 states. We report the run with the lowest resulting objective; Appendix~\ref{app:seed-variation} summarizes the results across all five initializations. As a numerical reference, we enumerate all possible nonempty supports and optimize the allocation conditional on each support. Appendix~\ref{app:planning-details} gives the numerical settings.

For both graph types, we study one randomly generated network. Both are composed of 12 nodes, use the same demand, generation, candidate sites, line properties, and operating-state variations; only the connections between buses differ. We scale $\kappa$ by $\Delta E$, the difference between the worst and best one-site mean energy expenditures within each instance. Appendix~\ref{app:planning-details} gives additional details.

\begin{figure}[t]
  \centering
  \includegraphics[width=\linewidth]{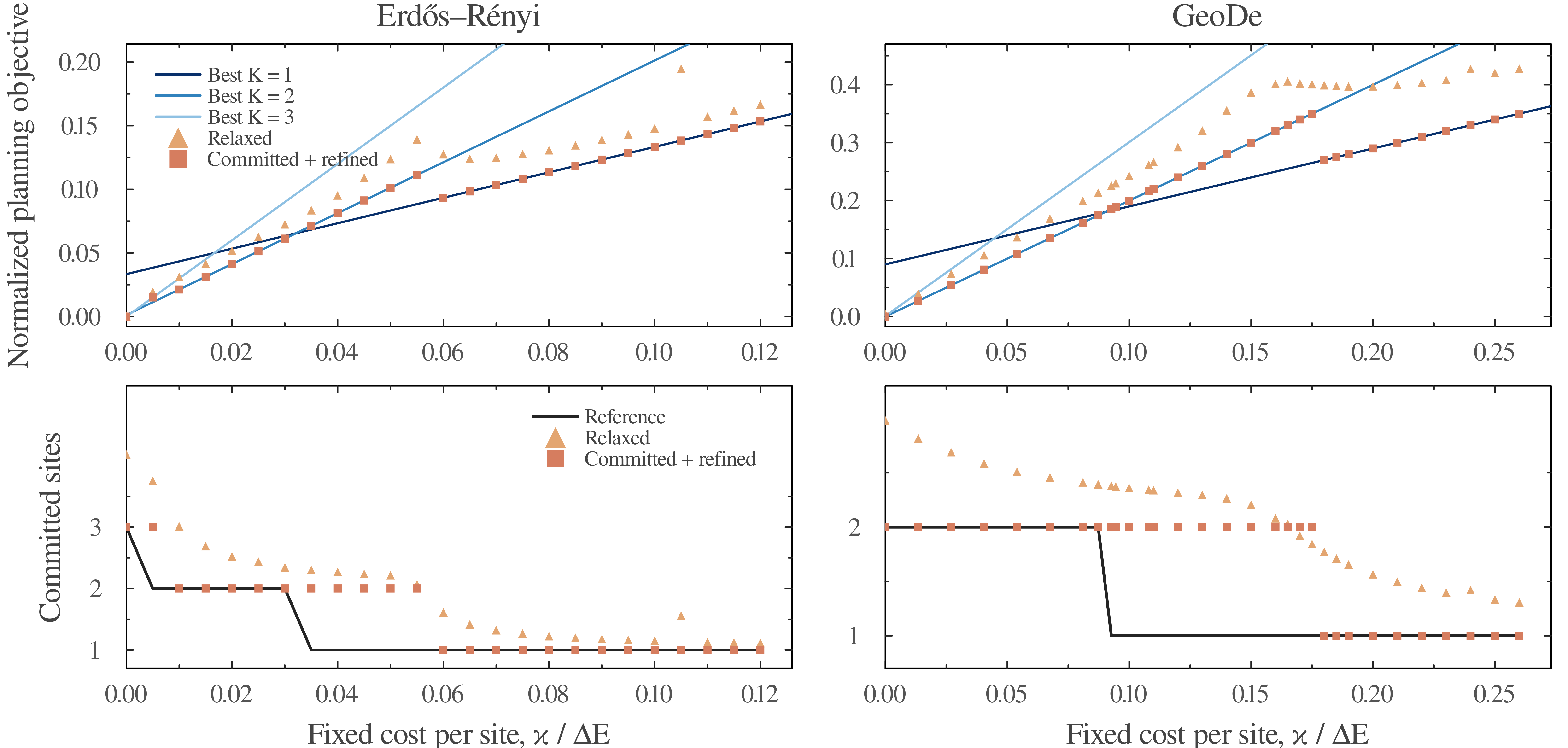}
  \caption{Portfolio recovery on the Erdős--Rényi and GeoDe
  instances. In the top row, solid lines show the numerical reference
  conditional on site count; squares show the committed and refined run with
  the lowest re-cleared objective among five initializations; and triangles
  show the raw relaxed result from that run. The bottom row
  compares reference, committed, and relaxed site counts. Each column subtracts
  its zero-penalty objective offset and divides by its one-site energy-cost
  spread, $\Delta E$.}
  \label{fig:portfolio-recovery}
\end{figure}

\subsection{Planning results}

Across both cost sweeps, the committed-and-refined objective closely tracks the numerical reference. The largest positive objective gap is $0.023\Delta E$ for Erd\H{o}s--R\'enyi and $0.085\Delta E$ for GeoDe; both occur where the learned procedure retains one additional site. Appendix~\ref{app:switch-lag} traces this delayed support switch to the smooth site-count relaxation itself rather than to an optimization failure.

Appendix~\ref{app:seed-variation} shows this tracking is not an artifact of reporting the best initialization: the mean re-cleared objective also remains close to the reference. Deviations and variation are largest near portfolio switches, where the site choice is most sensitive to the fixed cost.

\section{Discussion and conclusion}

These experiments separate two questions. Can market gradients identify where the load should be placed? In both networks, they reliably find the attractive buses and produce near-reference allocations.

Can a smooth relaxation determine exactly when a site should close? Here the answer is less satisfactory: near portfolio switches, the relaxation gradually shrinks a secondary site instead of closing it at the correct fixed cost.

This result is encouraging, but its scope is limited. Our experiments use two small synthetic instances, with six candidate sites and a limited set of operating states. They do not establish performance across networks or at realistic system scale. Larger problems will also make exhaustive enumeration impossible and may make initialization, local minima, and computational cost more important.

Future work should test the approach on larger power systems and historical market conditions. Another direction combines differentiable market clearing with explicit site-selection methods: gradients locate and size the load while a separate step selects the active sites.

\clearpage

\bibliographystyle{plainnat}
\bibliography{references}

\clearpage
\appendix

\section{Gradient-validation details}
\label{app:gradient-validation}

\paragraph{Directional checks.}
For one operating state $s$, let $f_s(z)=q(z)^\top\lambda_s(q(z))$, where $q(z)=D\operatorname{softmax}(z)$. We compare the reverse-mode directional derivative of $f_s$ along a centered unit vector $v$ with $[f_s(z+\epsilon v)-f_s(z-\epsilon v)]/(2\epsilon)$. For each of the Erd\H{o}s--R\'enyi and GeoDe families, we use topology seeds 1--3, load capacities 25, 50, and 100 MW, four of 12 generated operating states, three allocation seeds, three direction seeds, and $\epsilon\in\{10^{-3},10^{-4},10^{-5}\}$. This gives 972 checks per topology family.

We describe an active set by the lower, interior, or upper status of each line constraint and by the offer interval containing each generator dispatch. A finite-difference stencil is stable when its center and both endpoints have the same signature. All 1,944 directional stencils were stable. We treat directions whose analytic magnitude is below $10^{-3}$ as near zero and assess them by absolute rather than relative error. Grouping the remaining directions by topology, capacity, congestion regime, and $\epsilon$, the largest groupwise 90th-percentile relative error was $7.94\times10^{-5}$ for Erd\H{o}s--R\'enyi and $1.48\times10^{-4}$ for GeoDe. These checks validate the physical energy expenditure path; they intentionally exclude the straight-through derivative of the relaxed site count.

\paragraph{Active-set boundaries.}
We also scan 200-interval paths between concentrated candidate-site allocations, locate changes in the active-set signature by bisection, and test the appropriate derivative on each side. We request distances $10^{-2}$, $10^{-3}$, and $10^{-4}$ from the boundary and use a finite-difference step equal to 0.2 times that distance. When a stencil leaves its expected adjacent regime, we halve the distance down to a minimum of $10^{-4}$; if no valid stencil remains, we mark that side unresolved. We do not assert or test a derivative exactly at a boundary.

Each topology family produced 18 detected transitions. Table~\ref{tab:boundary-validation} reports the checks at requested distance $10^{-3}$. Unresolved transitions are excluded from the error summaries rather than counted as successful checks.

\begin{table}[h]
  \centering
  \caption{One-sided boundary checks at requested distance $10^{-3}$. Errors
  summarize informative resolved sides.}
  \label{tab:boundary-validation}
  \begin{tabular}{llrrrr}
    \hline
    Topology & Change & Found & Resolved & 90th percentile & Maximum \\
    \hline
    Erd\H{o}s--R\'enyi & line  & 10 & 8 & $2.44\times10^{-4}$ & $2.73\times10^{-4}$ \\
    Erd\H{o}s--R\'enyi & offer &  8 & 2 & $9.41\times10^{-6}$ & $9.83\times10^{-6}$ \\
    GeoDe              & line  &  8 & 7 & $5.61\times10^{-3}$ & $1.43\times10^{-2}$ \\
    GeoDe              & offer & 10 & 5 & $1.23\times10^{-1}$ & $3.34\times10^{-1}$ \\
    \hline
  \end{tabular}
\end{table}

The boundary study is therefore diagnostic rather than a blanket accuracy claim. In particular, a narrow GeoDe offer regime yields a maximum relative discrepancy of $0.334$ at the requested $10^{-3}$ distance, and solver noise becomes more prominent at $10^{-4}$. We retain these cases as limitations of finite-difference validation near active-set changes.

\section{Planning-experiment details}
\label{app:planning-details}

\subsection{Network construction and operating states}
\label{app:network-states}

Both cases place 50 MW of data-center load across six candidate buses. We first construct a common base system containing bus-level demand, generation offers, and a collection of line properties. We then connect the buses according to either the Erd\H{o}s--R\'enyi or GeoDe topology, using the original thermal ratings without rescaling. Finally, we generate 36 matched one-hour operating states by perturbing demand, offer prices, and available generation in the same way for both networks.

\newpage
\subsection{Market-clearing formulation}
\label{app:market-clearing-formulation}

Let $A$ be the oriented bus--line incidence matrix, let $B$ be the diagonal matrix of line susceptances, and let $M$ map the six candidate-site loads to their corresponding buses. For an allocation $q$, the nodal demand in operating state $s$ is $\widetilde d_s(q)=d_s+Mq$. We clear that state with the convex DC optimal-power-flow problem, which instantiates Equation~\eqref{eq:market-clearing} with $x=q$ and $y_s=(g_s,f_s,\phi_s)$:
\begin{equation}
  \begin{aligned}
    \underset{g_s,f_s,\phi_s}{\operatorname{minimize}}\quad
      & \sum_{j\in\mathcal G} C_{s,j}(g_{s,j}) \\
    \text{subject to}\quad
      & \widetilde d_{s,i}(q)+(Af_s)_i
        -\sum_{j:b(j)=i} g_{s,j}=0,
        && i\in\mathcal N, \\
      & f_s=BA^\top\phi_s,\qquad \phi_{s,r}=0, \\
      & 0\leq g_{s,j}\leq\overline g_{s,j},
        && j\in\mathcal G, \\
      & -\overline f_\ell\leq f_{s,\ell}\leq\overline f_\ell,
        && \ell\in\mathcal L,
  \end{aligned}
  \label{eq:concrete-market-clearing}
\end{equation}
where $g_s$ is generator dispatch, $f_s$ is line flow, $\phi_s$ is the vector of bus voltage angles, $b(j)$ is the bus containing generator $j$, and $r$ is the reference bus. The sets $\mathcal N$, $\mathcal G$, and $\mathcal L$ collect the buses, generators, and lines; the dispatch bounds and the flow--angle relation carve out the feasible set $\mathcal Y_s$ of Equation~\eqref{eq:market-clearing}, and the state $\omega_s$ supplies the demand $d_s$, the offer curves, and the available quantities $\overline g_s$. Each generator offer contains three quantity--marginal-price breakpoints $(\xi_{s,j,k},p_{s,j,k})$. If $\pi_{s,j}$ denotes their piecewise-linear interpolation, then the production cost used in Equation~\eqref{eq:concrete-market-clearing} is its integral,
\begin{equation}
  \begin{aligned}
    C_{s,j}(g)
      &=\int_0^g \pi_{s,j}(u)\,\mathrm du, \\
    \pi_{s,j}(u)
      &=p_{s,j,k}
        +\frac{p_{s,j,k+1}-p_{s,j,k}}
               {\xi_{s,j,k+1}-\xi_{s,j,k}}
         (u-\xi_{s,j,k}),
      && u\in[\xi_{s,j,k},\xi_{s,j,k+1}].
  \end{aligned}
  \label{eq:interpolated-offer-cost}
\end{equation}
Thus, the increasing interpolated marginal offers produce piecewise-quadratic, strictly convex generation costs. The dual variable of the nodal-balance constraint at bus $i$ is the locational marginal price (LMP) $\lambda_{s,i}(q)$ used in the planning objective. Within a fixed nondegenerate active-set regime, the market KKT system defines a locally differentiable primal--dual solution map, and the LMPs generally vary with the data-center allocation. Some directions can nevertheless have zero sensitivity---for example, reallocating a fixed total load in an uncongested state with a uniform nodal price---and a unique derivative need not exist exactly where an offer segment or transmission constraint changes status. Section~\ref{app:gradient-validation} evaluates both stable active-set regimes and one-sided behavior near such changes.

\subsection{Shift-factor implementation and LMP recovery}
\label{app:shift-factor-implementation}

The implementation (\texttt{DiffOpt.jl} over Ipopt) clears each state in an equivalent shift-factor form of Equation~\eqref{eq:concrete-market-clearing}~\citep{stott2009} rather than in the angle form above. Eliminating $f_s$ and $\phi_s$ through $f_s=BA^\top\phi_s$ aggregates the nodal balances into a single system-wide balance equality and expresses each line limit as a pair of inequalities on shift-factor-weighted net nodal injections, with the shift-factor matrix $H$ computed from $A$ and $B$ relative to the reference bus $r$. Every line is monitored in both directions at its original rating, so the two forms have identical solutions. For numerical robustness, the solved model softens these constraints with penalized slack variables, at penalty prices of 5000 (line limits) and 10000 (system balance) in the market's price units. The penalties cap the corresponding duals but sit far above every cleared price, and the recorded maximum constraint violation is zero across all reported runs, so the reported results coincide with the hard-constrained model of Equation~\eqref{eq:concrete-market-clearing}. Nodal prices are recovered from the duals of the shift-factor form as $\lambda_{s,i}=\sigma_s+\sum_{k}H_{ki}\,\mu_{s,k}$, where $\sigma_s$ is the dual of the system balance and $\mu_{s,k}$ are the duals of the line inequalities. This composition equals the nodal-balance duals of the angle form, and it is the map through which reverse-mode differentiation propagates price sensitivities, with the entries of $H$ entering the chain rule as constants.

\subsection{Objective normalization and optimization protocol}
\label{app:optimization-protocol}

We express the fixed site cost as $\kappa/\Delta E$, where $\Delta E$ is the difference between the worst and best one-site mean energy expenditures in that instance. This normalization measures construction cost relative to the energy savings available from choosing a better single site. The resulting dimensionless sweep spans $0$ to $0.12$ for Erd\H{o}s--R\'enyi and $0$ to $0.26$ for GeoDe. The scale factors are computed from the instances, rather than chosen: $\Delta E\approx735$ for Erd\H{o}s--R\'enyi and $\Delta E\approx566$ for GeoDe, in the objective's units.

The forward-pass temperature is $\tau=0.02$, while the straight-through backward-pass temperature is geometrically annealed from $2.0$ to $0.2$ over training. For each penalty, the learned method runs full-batch Adam for 450 iterations with learning rate 0.02. Its five initial logits are seeded Gaussian perturbations with standard deviation 0.01 around zero, and hence near the uniform allocation. We discard sites assigned less than 1 MW, renormalize the remaining allocation to 50 MW, and refine it on the committed support using two 50-step Adam runs followed by pairwise coordinate refinement. We then re-clear all 36 states with the exact number of retained sites and select the run with the lowest resulting objective. These five runs vary only the nearby optimizer initialization, not the network or market states.

\subsection{Numerical reference and validation}
\label{app:numerical-reference}

For the numerical reference, we enumerate all $2^6-1=63$ nonempty supports and numerically minimize mean energy expenditure conditional on each support, requiring every selected site to receive at least 1 MW. Supports of size at most three use a simplex grid with denominator 25; larger supports use four 150-step Adam runs with learning rate 0.03. Pairwise coordinate refinement starts at a 1 MW step and terminates at 0.02 MW or after 20 sweeps. We add $\kappa$ times the support size and take the lower envelope over supports. We refer to this lower envelope as the \emph{numerical optimum}. This is an operational term rather than a claim of solver-certified global optimality: the discrete support search is exhaustive, but the allocation conditional on each support is solved numerically.

Single-site values are exact. As an additional check, we independently probed all 15 two-site supports in each instance. Combining this search with the conditional optimizer produces no material change in the best two-site value: it finds no improvement for Erd\H{o}s--R\'enyi and an improvement of only $6.6\times10^{-5}$ objective units for GeoDe. The GeoDe reference envelope uses only one- and two-site portfolios, as does the Erd\H{o}s--R\'enyi envelope except at zero fixed cost, where it uses three sites. These checks provide strong evidence that the numerical optimum is close to the global optimum, while the three-site Erd\H{o}s--R\'enyi solution remains uncertified. The experiment records every support, optimizer seed, and training trace; parallel execution changes scheduling but not the seeded problem or per-run configuration.

\subsection{Initialization sensitivity}
\label{app:seed-variation}

\begin{figure}[htbp]
  \centering
  \includegraphics[width=0.95\linewidth]{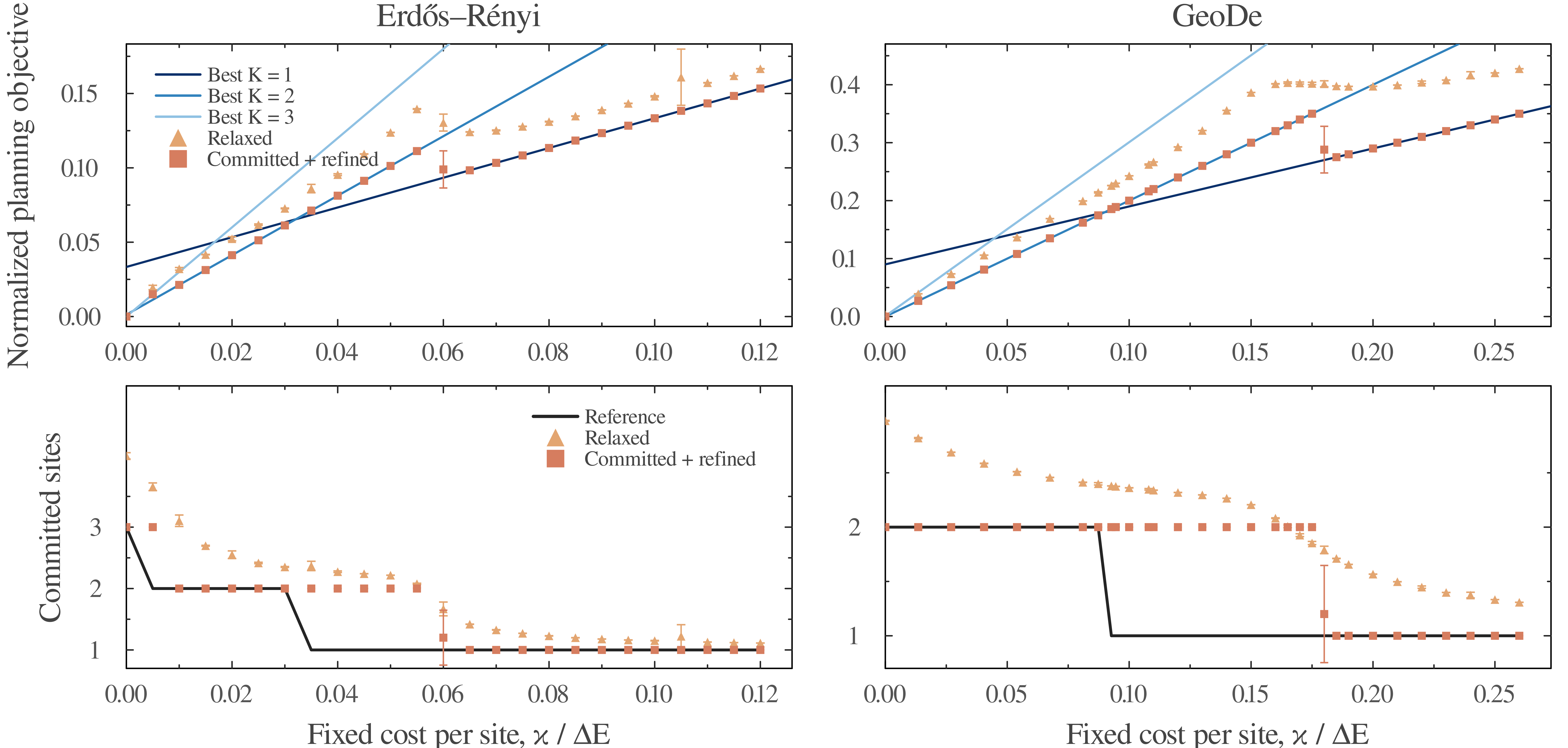}
  \caption{Mean $\pm$ one sample standard deviation across five seeded optimizer
  initializations. Squares aggregate the committed and refined objectives and
  site counts; triangles aggregate the relaxed objectives and smooth site
  counts. Away from the two sampled transition points, this aggregation
  preserves the qualitative pattern shown by the best-of-five curves.}
  \label{fig:portfolio-recovery-mean}
\end{figure}

Figure~\ref{fig:portfolio-recovery-mean} aggregates all five runs at each evaluated penalty instead of selecting the lowest-objective run. The error bars measure variation across the five tested optimizer initializations only: the network instance, market states, hyperparameters, commitment rule, and post-commitment refinement remain fixed. They are descriptive sample standard deviations, not confidence intervals over networks or operating states. Thus, this analysis characterizes local initialization stability around the near-uniform allocation; broader basin sensitivity would require a wider initialization study.

\section{Why the learned support switch lags the reference}
\label{app:switch-lag}

In Figure~\ref{fig:portfolio-recovery}, the learned portfolios keep a second site open at fixed costs where the numerical reference has already switched to a single site. This appendix shows that the lag is a measurable property of the relaxed objective the optimizer is given, not a failure to minimize it: training replaces the site count $\lVert q\rVert_0$ with the smooth count $K_\tau(w)$, so the relevant question is how much load the \emph{smoothed} objective assigns to a second site at each fixed cost.

\begin{figure}[htbp]
  \centering
  \includegraphics[width=0.95\linewidth]{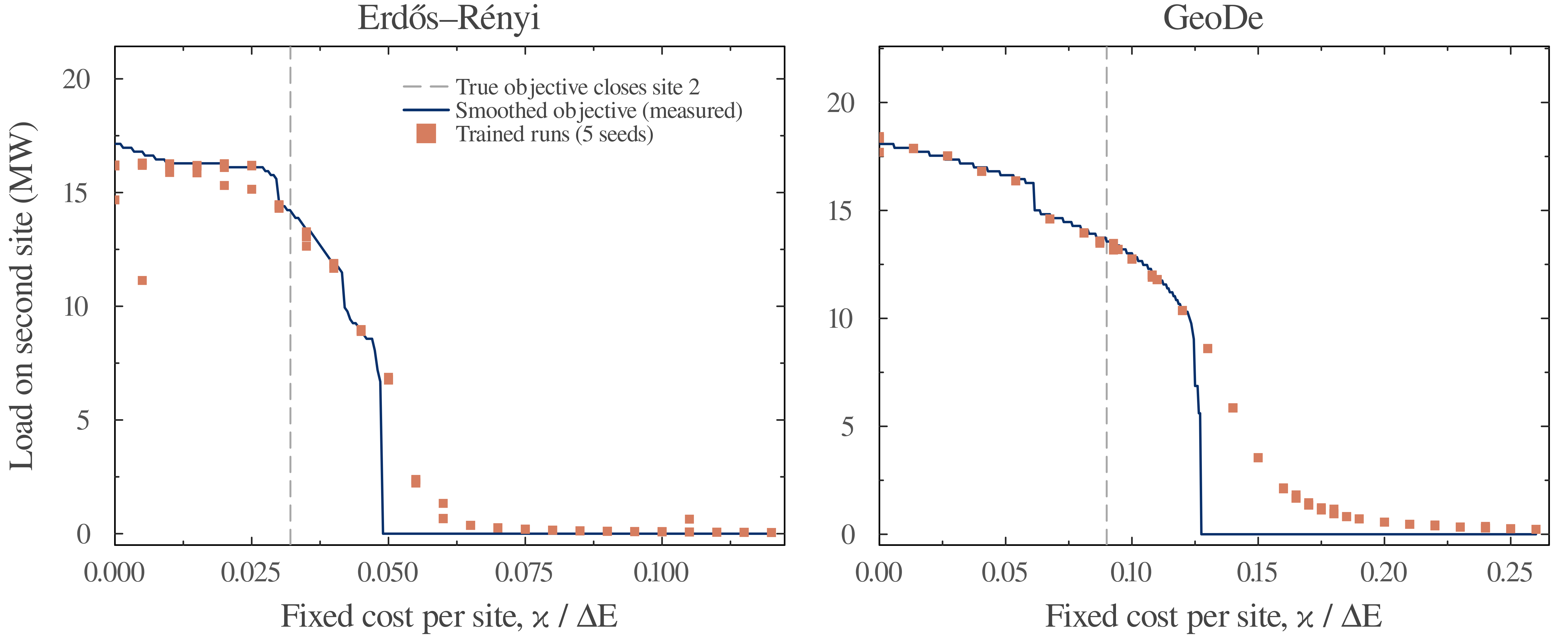}
  \caption{Second-site load requested by the smoothed training objective
  (solid line), measured by evaluating $E(w)+\kappa K_\tau(w)$ with forward
  market clears along the path between the best one-site and reference
  two-site allocations, compared with the relaxed allocations reached by the
  trained runs (squares; five initializations per fixed cost). The dashed
  line marks the fixed cost at which the true objective closes the second
  site. The smoothed objective shrinks the second site past this breakpoint
  instead of closing it, and the trained runs follow the measured curve.}
  \label{fig:switch-prediction}
\end{figure}

We answer this question by direct measurement, without any training. For each instance we interpolate linearly between the best single-site allocation and the reference two-site allocation, and evaluate the smoothed objective $E(w)+\kappa K_\tau(w)$ at every intermediate allocation, where $E(w)$ is the mean re-cleared energy expenditure obtained from forward market clears and $\tau=0.2$ is the terminal backward temperature of the annealing schedule. For each $\kappa$ we then follow this objective downhill from the two-site end of the path and record the second-site load at which descent stops. Figure~\ref{fig:switch-prediction} compares this prediction with the relaxed allocations that the trained runs actually reach.

Two mechanisms are visible. First, because $K_\tau$ charges a partially open site less than a fully open one, the smoothed objective responds to a rising fixed cost by shrinking the second site rather than closing it: its preferred second-site load decreases smoothly through the true-objective breakpoint and collapses to zero only at $\kappa/\Delta E\approx0.049$ on Erd\H{o}s--R\'enyi and $\approx0.13$ on GeoDe, versus breakpoints of $0.032$ and $0.090$. The trained runs track the measured curve closely throughout this region, which accounts for most of the observed lag. Second, near the collapse the smoothed objective is bistable: the shrunken two-site configuration and the collapsed one-site configuration are separated by a barrier, so gradient descent keeps whichever configuration it approaches from, and a collapsed iterate cannot reopen the second site. This bistability matches the disagreement between initializations observed exactly at the switch points (Appendix~\ref{app:seed-variation}). The short tail of small second-site loads beyond the predicted collapse, up to $\kappa/\Delta E\approx0.06$ and $\approx0.18$, reflects the finite annealing schedule: the penalty reaches its terminal temperature only late in training, and the remaining drain does not complete within the 450 iterations before commitment.

\end{document}